\documentclass[11pt]{article}
\usepackage{acl2016}
\usepackage{times}
\usepackage{latexsym}
\usepackage{verbatim}

\aclfinalcopy % Uncomment this line for the final submission
\usepackage{graphicx}
\usepackage{amsmath}
\usepackage{amsfonts}
\usepackage{multirow}
\usepackage{amssymb}
\usepackage{mathrsfs}
\usepackage{mathtools}
\usepackage{url}

\usepackage{pgfplots}
\usetikzlibrary{patterns}
\pgfplotsset{compat=1.5}

\usepackage{listings}
\makeatletter
\g@addto@macro\normalsize{%
  \setlength\abovedisplayskip{8pt}
  \setlength\belowdisplayskip{8pt}
  \setlength\abovedisplayshortskip{8pt}
  \setlength\belowdisplayshortskip{8pt}
}
\makeatother

\title{LSTM-based Mixture-of-Experts for Knowledge-Aware Dialogues}

\author{Phong Le \\ University of Amsterdam\\ p.le@uva.nl
\And Marc Dymetman, Jean-Michel Renders \\ 
Xerox Research Centre Europe \\
\{firstname.lastname\}@xrce.xerox.com}

\begin{document}

\maketitle

\begin{abstract}
We introduce an LSTM-based method for dynamically integrating several word-prediction experts 
to obtain a conditional language model which can be good simultaneously at several subtasks.
We illustrate this general approach with an application to dialogue where we integrate a neural chat model, 
good at conversational aspects, with a neural question-answering model, 
good at retrieving precise information from a knowledge-base, and show 
how the integration combines the strengths of the independent components.
We hope that this focused contribution will attract attention on the benefits of using such mixtures of experts
in NLP. 
\footnote{The work was performed during Phong Le's internship at XRCE in 2015.}
\end{abstract}

%%%%%%%%%%%%%%%%%%%%%%%%%%%%%%%%%%%%%%%%%%%%%%%%%%%%%%%%%%%%%%%%%%%
\section{Introduction}
\label{section intro}
%The traditional architecture for virtual agents in dialogue systems \cite{Jokinen2009} involves a combination of components: a \emph{Dialogue Manager} takes input from a \emph{Natural Language Understanding} process, updates an internal state, consults a \emph{Knowledge Base}, decides on the next action to take, and communicates this action to a \emph{Natural Language Generation} process through a \emph{Dialog Act} representation. These different processes require a lot of expertise in the different technologies, considerable development and implementation effort to adapt each component to a new domain, and are only partially trainable (if at all).

The traditional architecture for virtual agents in dialogue systems \cite{Jokinen2009} 
involves a combination of several components, 
which require a lot of expertise in the different technologies, 
considerable development and implementation effort to adapt each component to a new domain, 
and are only partially trainable (if at all).
Recently, \newcite{vinyals2015neural}, \newcite{serban2015hierarchical}, \newcite{Shang2015} proposed to replace this complex architecture by a single network (such as a Long Short Term Memory (LSTM) \cite{hochreiter1997long}) that predicts the agent's response from the dialogue history up to the point where it should be produced: this network can be seen as a form of conditional neural language model (LM), where the dialogue history provides the context for the production of the next agent's utterance. 

Despite several advantages over the traditional architecture (learnability, adaptability, better approximations to human utterances), this approach is inferior in one dimension: it assumes that all the knowledge required for the next agent's utterance has to be implicitly present in the dialogues over which the network is trained, and to then be precisely memorized by the network, while the traditional approach allows this knowledge to be dynamically accessed from external knowledge-base (KB) sources, with guaranteed accuracy.

To address this issue, we propose the following approach. As in \newcite{vinyals2015neural}, we first do train a conditional neural LM based on existing dialogues, which we call our \emph{chat model}; this model can be seen as an ``expert'' about the conversational patterns in the dialogue, but not about its knowledge-intensive aspects. Besides, we train another model, which this time is an expert about these knowledge aspects, which we call our \emph{QA model}, due to its connections to Question Answering (QA). We then combine these two expert models through an LSTM-based \emph{integration model}, which at each time step, encodes the whole history into a vector and then uses a softmax layer to compute a probability mixture over the two models, from which the next token is then sampled.

%% Although QA has been intensively studied recently, existing QA corpora and methods for
%% generating data (e.g., \newcite{fader2013paraphrase}) 
%% hardly meet our need here. 
%% This is because our case is very different from (and somewhat more difficult than) 
%% traditional QA set-ups in which questions are independent. 
%% In our case several scenarios are possible, resulting from the chat interaction 
%% (e.g., in a chat, questions can be related as in Figure~\ref{figure result chat example}).
%% We therefore propose a method generating artificial QA data that can cover several scenarios.

While here we combine in this way only two models, this \emph{core contribution} of our paper is immediately generalizable to several expert models, each competent on a specific task, where the (soft) choice between the models is done through the same kind of contextually-aware ``attention'' mechanism. Additional smaller contributions consist in the neural regime we adopt for training the QA model, and the way in which we reduce the memorization requirements on this model.
\footnote{Concurrently with our work, \newcite{DBLP:journals/corr/YinJLSLL15} 
propose a similar idea focussing only on QA in a traditional set-up.
Our case is more difficult because of the chat interaction; 
and the integration framework we propose is generally applicable to situations 
where a pool of word-prediction ``experts'' compete for attention during the generation of text.}

%%%%%%%%%%%%%%%%%%%%%%%%%%%%%%%%%%%%%%%%%%%%%%%%%%%%%%%%%%%%%%%%%%%
%\section{LSTM-based Mixture of Experts for Language Modelling}
\section{LSTM-based Mixture of Experts}
\label{section lstm moe}

\begin{figure}
\centering
\includegraphics[width=.4\textwidth]{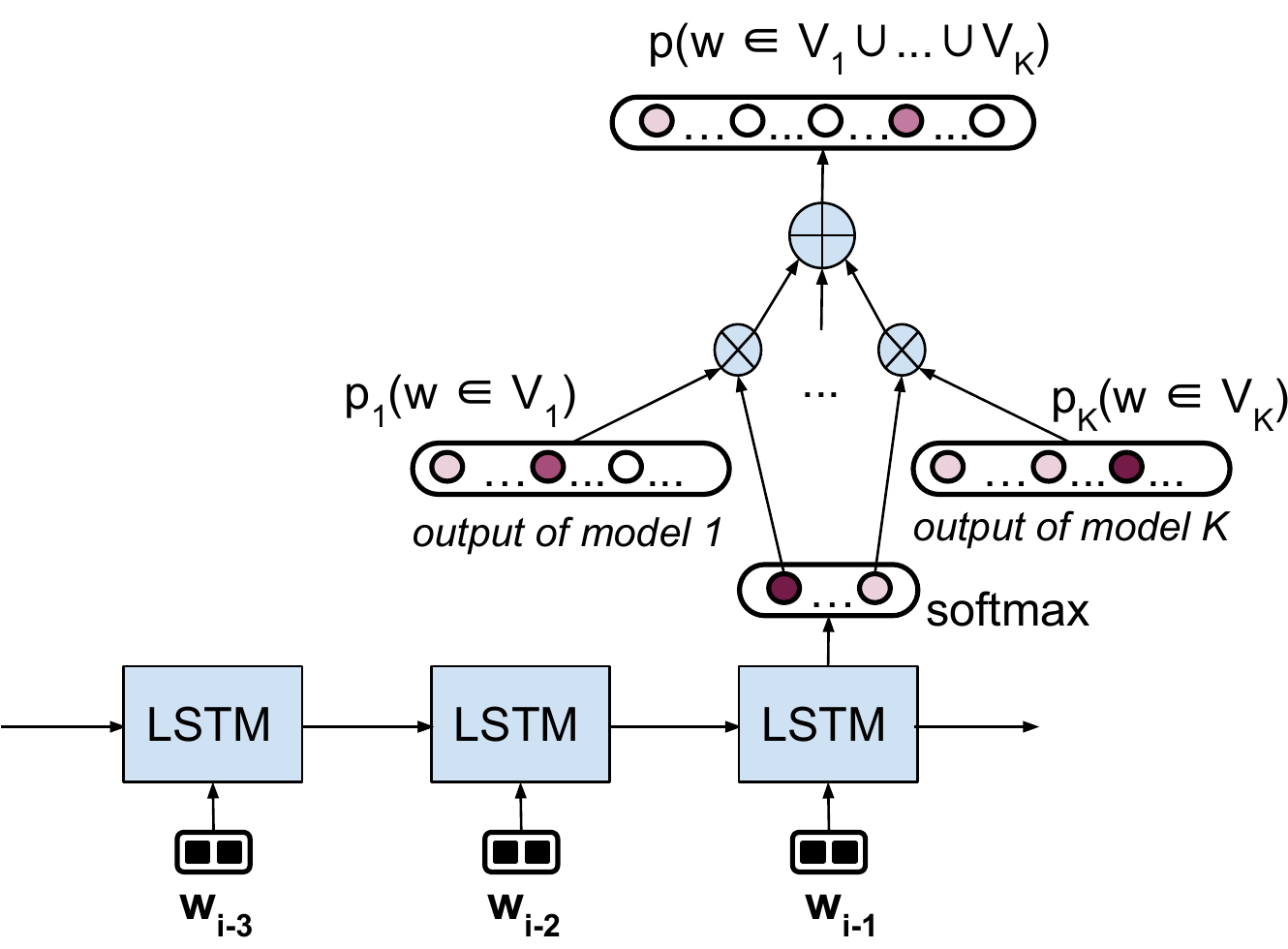}
\caption{LSTM-based mixture-of-experts for Language modelling. $\otimes$ denotes 
multiplication, $\oplus$ denotes sum.}
\label{fig lstm moe}
\end{figure}

The method is illustrated in Figure~\ref{fig lstm moe}. 
Let $w_1^t = w_1...w_t$ be a history over words. 
We suppose that we have $K$ models each of which can compute a distribution over its own vocabulary $V_k$ : 
$p_k(w \in V_k | w_1^t)$, for  $k\in [1,K]$. 
We use an LSTM to encode the history word-by-word into a vector $\mathbf{h}_t$ which is the hidden state 
of the LSTM at time step $t$. We then 
use a softmax layer to compute the probabilities
$p(k|w_1^t) = \frac{e^{u(k,\mathbf{h}_t)}}{\sum_{k'=1}^K e^{u(k',\mathbf{h}_t)}}$
where $\left[ u(1,\mathbf{h}_t),...,u(K,\mathbf{h}_t) \right]^T
= \mathbf{W} \mathbf{h}_t + \mathbf{b}$,
$\mathbf{W} \in \mathbb{R}^{K \times dim(\mathbf{h}_t)}, \mathbf{b}\in \mathbb{R}^{K}$. 
The final probability of the next word is then:
\begin{equation}
\label{equation lstm moe}
p(w | w_1^t) = \sum_{k=1}^K p(k | w_1^t)\: p_k(w | w_1^t) .
\end{equation}

Our proposal can be seen as bringing together two previous lines of research within an LSTM framework. Similar to the \emph{mixture-of-experts} technique of \newcite{jacobs1991task}, we predict a label by using a ``gating'' neural network to mix the predictions of different experts based on the current situation, and similar to the approach of \newcite{florian1999dynamic}, we dynamically combine distributions on words to produce an integrated LM.\footnote{\newcite{florian1999dynamic} focus on the combination of topic-dependent LMs, while in our case, the components can be arbitrary distributions over words --- we later use a component that produces answers to questions appearing in the text.} In our case, the labels are words, the gating neural network is an LSTM that stores a representation of a long textual prefix, and the combination mechanism is trained by optimizing the parameters of this LSTM.

%%%%%%%%%%%%%%%%%%%%%%%%%%%%%%%%%%%%%%%%%%%%%%%%%%%%%%%%%%%%%%%%%%
\section{Data}
\label{section setup}

\begin{figure}
\centering
\includegraphics[width=0.5\textwidth]{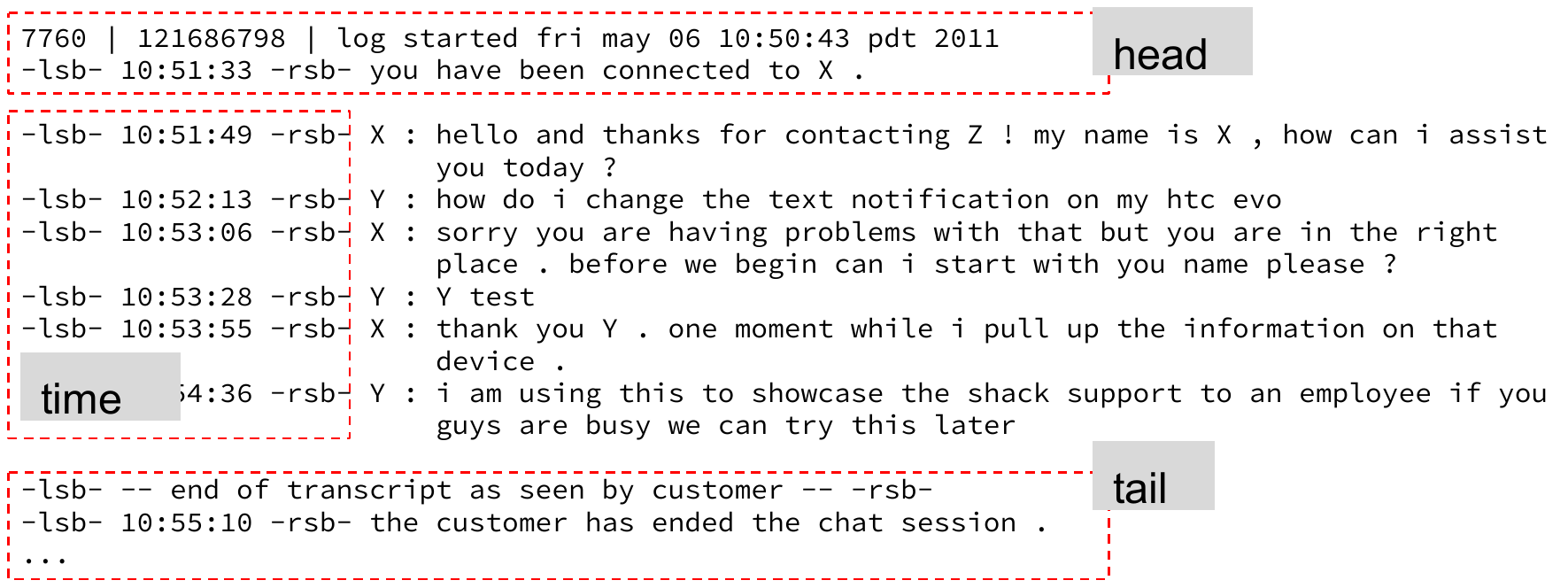}
\caption{An example dialogue.}
\label{fig dialog}
\end{figure}

Our corpus consists of 165k dialogues from a ``tech company''
in the domain of mobile telephony support. 
We split them into train, development, and test sets whose sizes are 
145k, 10k, and 10k dialogues. 
We then tokenize, lowercase each dialogue, and 
remove unused information such as head, tail, chat time 
(Figure~\ref{fig dialog}).
For each response utterance 
found in a dialogue, we create a context-response pair whose context consists of 
all sentences appearing before the response. 
This process gives us 973k/74k/75k pairs for 
training/development/testing.

\paragraph{Knowledge-base}
The KB we use in this work consists of 1,745k
device-attribute-value triples, e.g., 
(Apple iPhone 5; camera megapixels; 8.0).
There are 4729 devices and 608 attributes. 
Because we consider only numeric values, only triples that 
have numeric attributes
are chosen, resulting in a set of 65k triples 
of 34 attributes.

%Since this KB uses standard units (e.g., ``megabyte'' for 
%storage, ``mhz'' for CPU frequency) but agents sometimes use 
%other units (e.g., ``gigabyte'', ``ghz''), we convert each value in 
%the KB to other units if applicable. 

\paragraph{Device-specification context-response pairs}
\label{subsection target pairs}
Our target context-response pairs are those in which the client 
asks about numeric value attributes. 
We employ a simple heuristic to select 
target context-response pairs: a context-response pair 
is chosen if its response contains a number and one of the following 
keywords: cpu, processor, ghz, mhz, memory, mb(s),
gb(s), byte, pixel, height, width, weigh, 
size, camera, mp, hour(s), mah. 
Using this heuristic, we collect 17.6k/1.3k/1.4k pairs for 
training/dev/testing. 
These sets are significantly smaller than those extracted 
above. 
%Besides, only  6.8\% value tokens are found in the KB.   

%To identify how good are chosen pairs, we randomly pick up 
%97 pairs and examine them. The result is given in 
%Table~\ref{table pair-stats}. One can see that the employed 
%heuristic is not adequate. But we will show later that, 
%even being trained on such ``dirty'' data, our model can perform 
%well. 

%\begin{table}[!ht]
%\centering
%\begin{tabular}{p{7cm}|r}
 %good & 36.1 \% \\ \hline
 %device name is not in the KB & 4.1 \% \\ \hline
 %values given by the response are not the same with ones given by the KB & 11.3 \% \\ \hline
 %the context doesn't contain any device name & 2.1 \% \\ \hline
 %the context contains more than 1 device names & 13.4 \% \\ \hline
 %the context contains more than 1 attributes & 3.1 \% \\ \hline
 %irrelevant to the task & 23.7 \% \\ \hline
 %device name is not recognizable & 4.1 \% \\ \hline
 %values are not recognizable & 2.1 \% 
%\end{tabular}
%\caption{97 chosen pairs are examined.}
%\label{table pair-stats}
%\end{table}

%%%%%%%%%%%%%%%%%%%%%%%%%%%%%%%%%%%%%%%%%%%%%%%%%%%%%%%%%%%%%%%%%%%
\section{KB-aware Chat Model}

\subsection{Neural Chat Model}
\label{section neural chat}

Ouur corpus is comparable to the one described in \newcite{vinyals2015neural}'s 
first experiment, and we use here a similar neural chat model. 

Without going into the details of this model for lack of space, this model uses a LSTM to encode into a vector the sequence of words observed in a dialogue up to a certain point, and then this vector is used by another LSTM for generating the next utterance also word-by-word. The approach is reminiscent of seq2seq models for machine translation such as \cite{sutskever2014sequence}, where the role of ``source sentence'' is played by the dialogue prefix, and that of ``target sentence'' by the response utterance.

%%%%%%%%%%%%%%%%%%%%%%%%%%%%%%%%%%%%%%%%%%%%%%%%%%%%%%%%%%%%%%%%%%%
\subsection{Neural Question Answering Model}
\label{section qa}

In a standard setting, a question to query a KB 
must be formal (e.g., SQL). 
However, because a human-like QA system should take 
natural questions as input, % instead of formal expressions. 
we build a neural model to translate natural questions to 
formal queries. 
This model employs an LSTM to encode a natural question into a vector. 
It then uses two softmax layers
to predict the device name and the attribute.
This model is adequate here, since we focus on the QA situation where 
the client asks about device specifications.
For more complex cases, more advanced QA models should 
be considered (e.g., \newcite{bordes-chopra-weston:2014:EMNLP2014},
\newcite{yih-EtAl:2015:ACL-IJCNLP}). 

Given question $w_1^l$, 
the two softmax layers give 
us a distribution over devices $p_{d}(\bullet | w_1^l)$ and 
a distribution over attributes $p_{a}(\bullet | w_1^l)$. 
Using the KB, we can compute a distribution over the set 
$V_{qa}$ of all values found in the KB, by marginalizing over $d,a$:
\begin{equation}
p_{qa}(v | w_1^l) = \sum_{\langle d,a,v \rangle \in T} p_{d}(d | w_1^l) p_{a}(a | w_1^l),
\label{equation neural qa distr}
\end{equation}
where $T$ is the set of all triples in the KB.  Initial experiments
showed that predicting values in this indirect way significantly
improves the accuracy compared to employing a single softmax layer to
predict values directly, because it minimizes the memorization
requirements on the hidden states.

\paragraph{Data Generation}
%One difficulty is that we do not have natural questions to train the
%QA model, and existing QA methods for generating data (e.g.,
%\newcite{fader2013paraphrase}) are almost unusable.  We thus propose a
%generation process.\footnote{Although QA has been intensively studied
%  recently, existing QA corpora and methods for generating data (e.g.,
%  \newcite{fader2013paraphrase}) hardly meet our need here.  This is
%  because our case is very different from (and somewhat more difficult
%  than) traditional QA set-ups in which questions are independent.  In
%  our case several scenarios are possible, resulting from the chat
%  interaction (e.g., in a chat, questions can be related as in
%  Figure~\ref{figure result chat example}).  We therefore propose a
%  method generating artificial QA data that can cover several
%  scenarios. }
Although QA has been intensively studied
  recently, existing QA corpora and methods for generating data (e.g.,
  \newcite{fader2013paraphrase}) hardly meet our need here.  This is
  because our case is very different from (and somewhat more difficult
  than) traditional QA set-ups in which questions are independent.  In
  our case several scenarios are possible, resulting from the chat
  interaction (e.g., in a chat, questions can be related as in
  Figure~\ref{figure result chat example}).  We therefore propose a
  method generating artificial QA data that can cover several
  scenarios.

For each tuple $<$device name, attribute$>$, we paraphrase the device
name by randomly dropping some words (e.g., ``apple iphone 4'' becomes
``iphone 4''), and paraphrase the attribute using a small handcrafted
dictionary and also randomly dropping some words (``battery talk
time'' becomes ``battery life'' which can become ``battery'').  We
then draw a sequence of $l$ words from a vocabulary w.r.t word
frequency, where $l \sim Gamma(k,n)$ (e.g., ``i what have''), and
shuffle these words.  The output of the final step is used as a
training datapoint like: {\ttfamily have \textit{iphone 4} what
  \textit{battery} i $\rightarrow$ apple\_iphone\_4
  battery\_talk\_time}.  To make it more realistic, we also generate
complex questions by concatenating two simple ones.  Such questions
are used to cover the dialogue scenario where the client continues
asking about another device and attribute.  In this case, the system
should focus on the latest device and attribute.

Using this method, we generate a training set of 7.6m datapoints
and a development set of 10k.

%%%%%%%%%%%%%%%%%%%%%%%%%%%%%%%%%%%%%%%%%%%%%%%%%%%%%%%%%%%%%%%%%%%
\subsection{Integration}

We now show how we integrate the chat model with the QA model
using the LSTM-based mixture-of-experts method. 
The intuition is the following: the chat model is in charge of generating 
smooth responses into which the QA model ``inserts'' values retrieved from 
the KB. 
Ideally, we should employ an independent LSTM for the purpose of computing 
mixture weights, as in Section~\ref{section lstm moe}. 
However, due to the lack of training data, 
our integration model makes use of the chat model's hidden state 
to compute these weights. 
Because this hidden state captures 
the uncertainty of generating the next word, it is also able to detect 
whether or not the next word should be generated by the chat model. 

It is easy to see that the chat model is the backbone because 
most tokens should be generated by it.
The QA model, on the other hand, is crucial since 
we want the system to generate correct values. 
(E.g., the chat model alone cannot perform the chat 
shown in Figure~\ref{figure result chat example} precisely.) 
More importantly, in the future when new devices are released, 
we do not need to collect new chat data, 
which are often expensive, to retrain the chat model.

Let $C$ and $w_1^t$ be a context and words generated up to this point.
$p_c(\bullet | w_1^t, C)$ and $p_{qa}(\bullet | w_1^t, C)$ are given by 
the chat model and the QA model. 
% Following (\ref{equation lstm moe}), 
We then compute the distribution $p(\bullet | w_1^t, C)$ over $V_c \cup V_{qa}$ as 
a mixture of $p_c$ and $p_{qa}$:
\begin{equation*}
p(w | w_1^t,C) = \alpha . p_c(w | w_1^t,C ) + (1-\alpha) . p_{qa}(w | w_1^t,C)
\end{equation*}
where $\alpha = \sigma(\mathbf{w}^T \mathbf{h}_t^c + b) $, 
$\mathbf{h}_t^c$ is the hidden state of the chat model, 
$\sigma$ is the sigmoid function; $\mathbf{w} \in \mathbb{R}^{dim(\mathbf{h}_t^c)}$
and $b \in \mathbb{R}$. Note that the sigmoid is equivalent to the softmax 
for two output units.

\paragraph{Training}
To train this integration model, we keep the chat model and the QA model frozen, 
and minimize the objective:
\begin{equation*}
J(\theta) = - \sum_{(C,w_1^l) \in D} \sum_{t=0}^{l-1} 
\begin{matrix}
\beta(w_{t+1}).\\
\log  p(w_{t+1} | w_1^t, C;\theta)
\end{matrix}
+ \frac{\lambda}{2} ||\theta||^2
\end{equation*}
w.r.t. $\theta = (\mathbf{w}, b)$, 
where $\beta(w)=100$ if $w\in V_{qa} \setminus V_c$, $\beta(w)=1$ otherwise.
$\lambda$ is the regularization parameter and
$D$ is the training set.   
We set $\beta(w \in V_{qa} \setminus V_c)$ high because 
we want the training phase to focus on those tokens representing values in the 
KB but not supported by the chat model.

\paragraph{Decoding}
To find the most probable responses, our decoder 
employs the uniform-cost-search algorithm \cite{Russell:2003:AIM:773294}, 
which is guaranteed to find optimal solutions.  
We stipulate a constraint that a response is to answer not more than one question.

%%%%%%%%%%%%%%%%%%%%%%%%%%%%%%%%%%%%%%%%%%%%%%%%%%%%%%%%%%%%%%%%%%%
\section{Experiments}

We implement our models in C++ using CUDA Toolkit. 
Since evaluating a conversation system is still challenging, we, following \newcite{vinyals2015neural},
use word perplexity only. 
In our experiments, every LSTM has 1024 hidden units and 1024 memory cells. 
The vocabulary of the chat model has 19.3k words, that of the QA model 12.7k words. 

We firstly train the chat model on all chat data %(Section~\ref{section setup})
with the learning rate 0.01, and continue training it on the 
device-specification training data with a smaller learning rate, 0.001.
Using this smaller learning rate we expect that the model will not forget what it 
has learnt on all the chat corpus. Next, we train the QA model on 
the data generated in Section~\ref{section qa} with the learning rate 0.01.
Finally, we train the integration model on the device-specification training data
also with the learning rate 0.01.
 
\begin{figure}
\centering
\includegraphics[width=0.5\textwidth]{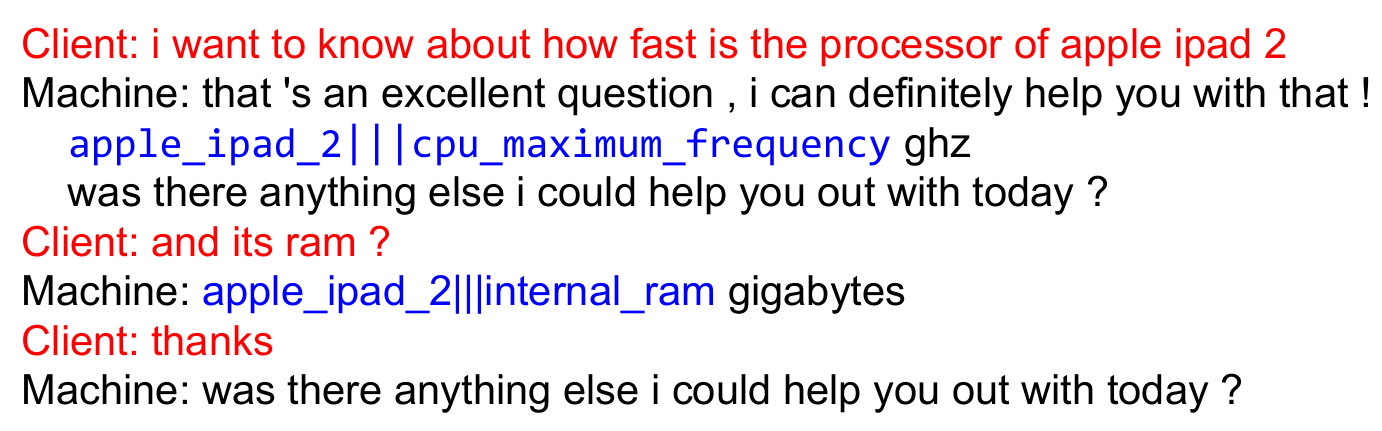}
\caption{A chat with the integration model.}
\label{figure result chat example}
\end{figure}

We report here initial results. Although the integration does not 
help  to decrease the perplexity on all tokens (15.4, compared to 14.7 
of the chat model),
%\footnote{The increase is due to the fact that the average value of $\alpha$ when the next token should be generated by the chat model alone 
% is $0.954 < 1$. That is the integration model wrongly puts weight 4.6\% of the time on words in $V_{qa}$, when it should select a word in $V_c$, which increases its perplexity in such cases.}
% 
it does help to decrease perplexity 38\%
on value tokens (46.8, compared to 75.8 of the chat model).
This shows that using the QA model is helpful, 
especially for predicting the right values, 
which are informationally more critical to the user than general words.\footnote{Perplexity is here a weak proxy for what a human evaluation of usefulness would provide. In terms of perplexity, the gain over value tokens does not help overall because value tokens are rare, only about 6.7\% of the data.
}
Figure~\ref{figure result chat example} shows a chat example between us 
and our integration model. To examine whether the model retrieves 
precise information, outputs are only shown here in the format $device|||attribute$.

\begin{comment}

\begin{table}
\centering
\begin{tabular}{c|c|c}
 & neural chat & integration \\ \hline
 all tokens & \textbf{14.7} & 15.4 \\
 only value tokens & 75.8 & \textbf{46.8} \\ 
\end{tabular}
\caption{Perplexities of without and with the integration.}
\label{table ppl neural chat vs integration}
\end{table}

\end{comment}

%%%%%%%%%%%%%%%%%%%%%%%%%%%%%%%%%%%%%%%%%%%%%%%%%%%%%%%%%%%%%%%%%%%
%\vspace*{-1mm}
\section{Conclusions}
%\vspace*{-1mm}

This short paper introduces a general LSTM-based mixture-of-experts method
for language modelling and illustrates the approach by integrating a
neural chat model with a neural QA model. The experimental results,
while limited to measures of perplexity, do show that the integration
model is capable of handling chats inside of which the user may ask
about device specifications; a more thorough and convincing evaluation
would require human assesments of the quality of the produced
responses.

%% On the other hand, there is room for improvement. 
%% For instance, we could employ a device recognizer (e.g., 
%% \newcite{radford-carreras-henderson:2015:EMNLP})
%% to help the QA model detect device names more accurately.
%% We could also generate artificial QA data that cover 
%% more chat scenarios.

We believe that the proposed integration method has potential for a wide range of applications. It allows to pool a number of different language models, each expert in a specific domain or class of problems (possibly trained independently based on the most appropriate data) and to generate the next word based on a competition between these models, under the supervision of an LSTM-based attention mechanism.

\bibliographystyle{acl2016}
\bibliography{ref}

\end{document}

\maketitle

%%%%%%%%%%%%%%%%%%%%%%%%%%%%%%%%%%%%%%%%%%%%%%%%%%%%%%%%%%%%%%%%%%%%%%%%%

\section{Extracting Context-response Pairs}

\paragraph{Extracting pairs} 
For each response 
found in a dialog, we create a context-response pair whose context consists of 
all sentences appearing before the response. 
For each context, to mark where its ends, we add to its tail a 
token ``$<$EOC$>$''. We use tokens ``$<$CLIENT$>$'' and ``$<$AGENT$>$''
to inform whether the following utterance is from the client 
or from the agent. 
For each response we add a token ``$<$EOR$>$'' to its tail. Below is an example: 
\begin{itemize}
\item context: {\footnotesize\ttfamily
$<$CLIENT$>$ how do i change the text notification on my htc evo
$<$AGENT$>$ sorry you are having problems with that but you are in the right place . 
before we begin can i start with you name please ?
$<$CLIENT$>$ Y test $<$EOC$>$
}
\item response: {\footnotesize\ttfamily
thank you Y . one moment while i pull up the information on that device .
$<$EOR$>$
}
\end{itemize}

\paragraph{Dealing with long turns} A turn could consist of more than one 
response, such as
\begin{lstlisting}
[...]
CLIENT: ok , so thats it ?
AGENT: that 's it !
AGENT: there may be some warranty i am checking
CLIENT: ok i will test it and follow up thursday
[...]
\end{lstlisting}
To inform that there should be another response following the current one
(e.g., ``there may be some warranty i am checking'' follows 
``that 's it !''), we add to the tail of the current response 
a token ``$<$pause$>$'' as follows
\begin{itemize}
\item context: {\footnotesize\ttfamily
[...] $<$CLIENT$>$ ok , so thats it ? $<$EOC$>$
}
\item response: {\footnotesize\ttfamily
that 's it ! $<$pause$>$
$<$EOR$>$
}
\end{itemize}
and the next context-response pair is
\begin{itemize}
\item context: {\footnotesize\ttfamily
[...] $<$CLIENT$>$ ok , so thats it ? $<$AGENT$>$  that 's it ! $<$pause$>$ $<$EOC$>$
}
\item response: {\footnotesize\ttfamily
there may be some warranty i am checking
$<$EOR$>$
}
\end{itemize}

\paragraph{Dealing with long responses} A response could be long, for 
instance when the agent copy a text from a document
\begin{lstlisting}
AGENT: step 1 :

a master reset will result in the loss of all data stored on the device including 3rd party applications .

this type of reset should only be used as a last resort after all other troubleshooting steps have been exhausted . it is highly recommended that you complete a backup of the device information before proceeding with the steps in this tutorial .

ensure the handset is turned off .

note : if the handset is frozen , remove and reinsert the battery to power it off .
step 2 :

press and hold the volume down key .
step 3 :

[...]
\end{lstlisting}
Because long responses make both training and generation difficult, 
we split a long response into smaller ones and use tokens ``$<$newline$>$''
to mark the break-points. 

Note that, during decoding (i.e., generating the best response), 
if the current output is ended with a token ``$<$newline$>$'' or ``$<$pause$>$'', 
the decoder will continue generating another response.

%%%%%%%%%%%%%%%%%%%%%%%%%%%%%%%%%%%%%%%%%%%%%%%%%%%%%%%%%%%%%%%%%%%%%%%%%

\section{List of Attributes}

\begin{lstlisting}
battery capacity
battery standby time
battery talk time
browser screen size
camera digital zoom factor
camera maximum resolution
camera megapixels
camera optical zoom factor
cpu
cpu maximum frequency
internal ram
internal storage
java app usable screen size
java max memory size
native app usable screen size
primary screen physical height
primary screen physical width
primary screen rotate
primary screen type
removable memory maximum size
rendering screen size
screen orientation
screen size
screen size char
secondary camera maximum resolution
secondary camera megapixels
secondary screen physical height
secondary screen physical width
secondary screen size
secondary screen size char
secondary screen type
sync contacts to removable memory
wallpaper external screen usable size
wallpaper internal screen usable size
\end{lstlisting}

%%%%%%%%%%%%%%%%%%%%%%%%%%%%%%%%%%%%%%%%%%%%%%%%%%%%%%%%%%%%%%%%%%%%%%%%%

\section{Paraphrase}

We paraphrase attributes using Table~\ref{tab paraphrase attr}.

\begin{table*}
\centering
\begin{tabular}{p{5cm}|p{8cm}}
 attribute & alternatives \\ 
 \hline \hline
 
battery capacity & battery size \\ \hline
battery standby time & battery life \\ \hline
battery talk time & battery life  \\ \hline
camera megapixels & megapixel, mega pixel, mp, mega pixels  \\ \hline
cpu maximum frequency & cpu, processor, power \\ \hline
internal ram & ram \\ \hline
internal storage & memory \\ \hline
primary screen physical height & screen height \\ \hline
primary screen physical width & screen width \\ \hline
removable memory maximum size & external memory, external storage, memory card, sd card \\ \hline
secondary camera maximum resolution & front camera resolution \\ \hline
secondary camera megapixels & front megapixel, front mega pixel, front mp, front mega pixels
\end{tabular}
\caption{Paraphrasing attributes. }
\label{tab paraphrase attr}
\end{table*}

%%%%%%%%%%%%%%%%%%%%%%%%%%%%%%%%%%%%%%%%%%%%%%%%%%%%%%%%%%%%%%%%%%%%%%%%%

\section{Visualizing Outputs of the KB-aware Model}

We choose two context-response pairs in the development set and visualize 
$\alpha$, $p_c$, $p_{qa}$. 
In Figure~\ref{fig alpha 1},  
we can see that $\alpha$ tends to have very high values ($>0.9$) for tokens 
(e.g., ``hours, days'') that should be generated by the neural chat model. 
On the other hand, it has lower values for tokens (e.g., ``220'') that should 
be generated by the QA model. Note that because the value ``9.2''  is not 
in the knowledge base, its $p_{qa}$ is zero. On the other hand, because 
``9.2'' often appears after ``220'' in the training data, its $p_{c}$
is high. 

\begin{figure}
\centering
\begin{tikzpicture}
\begin{axis}[
	xscale=1,
    ybar,
    enlargelimits=0.05,
    legend style={at={(0.45,-0.35)},
      anchor=north,legend columns=-1},
    symbolic x coords={standby, :, 220, hours, max, ., -lrb-, 9.2, days, -rrb-, $<$EOR$>$},
    xtick=data,
    bar width=3pt,
    xticklabel style={rotate=60},
    ]
\addplot[ybar, pattern=north east lines] coordinates {(standby,0.86) (:,0.99) (220,0.51) (hours,0.98) (max,1) (.,1) (-lrb-,1) (9.2,0.86) (days,1) (-rrb-,1) ($<$EOR$>$,1)};
\addplot coordinates {(standby,0.97) (:,0.98) (220,0.03) (hours,1) (max,0.99) (.,0.97) (-lrb-,1) (9.2,0.7) (days,1) (-rrb-,1) ($<$EOR$>$,0.59)};
\addplot[ybar, color=black] coordinates {(standby,0) (:,0) (220,0.42) (hours,0) (max,0) (.,0) (-lrb-,0) (9.2,0) (days,0) (-rrb-,0) ($<$EOR$>$,0)};
\legend{$\alpha$, $p_c$, $p_{qa}$}
\end{axis}
\end{tikzpicture}
\caption{ $\alpha$, $p_c$ and $p_{qa}$ for the context-response 
\{{\ttfamily
<CLIENT> what is the batt life on the new photon q <AGENT> that 's an excellent question , i can definitely help you with that ! <pause> <AGENT> talk : 7.5 hours max . -lrb- 450 minutes -rrb- <newline> <EOC>}\} - 
\{{\ttfamily 
standby : 220 hours max . -lrb- 9.2 days -rrb- <EOR>}\}. }
\label{fig alpha 1} 
\end{figure}

Figure~\ref{fig alpha 2} shows that neural QA model is not always helpful. In this 
example, the token ``2'' receives very high values of $\alpha$ and $p_c$. 
This is because the constituent ``plus 2'' appears many times in the training data. 
The neural chat model thus be able to remember this fact.  Therefore, 
the integration model chooses the neural chat model instead of the neural QA model 
to generate this token. 

\begin{figure}
\centering
\begin{tikzpicture}
\begin{axis}[
	xscale=1,
    ybar,
    enlargelimits=0.1,
    legend style={at={(0.45,-0.35)},
      anchor=north,legend columns=-1},
    symbolic x coords={plus, 2, gigabyte, card, included, $<$pause$>$, $<$EOR$>$},
    xtick=data,
    bar width=4pt,
    xticklabel style={rotate=60},
    ]
\addplot[ybar, pattern=north east lines] coordinates {(plus,0.93) (2,0.98) (gigabyte,1) (card,1) (included,1) ($<$pause$>$,1) ($<$EOR$>$,1)};
\addplot coordinates {(plus,0.9) (2,0.53) (gigabyte,0.96) (card,0.82) (included,0.91) ($<$pause$>$,0.48) ($<$EOR$>$,1)};
\addplot[ybar, color=black] coordinates {(plus,0) (2,0.06) (gigabyte,0) (card,0) (included,0) ($<$pause$>$,0) ($<$EOR$>$,0)};
\legend{$\alpha$, $p_c$, $p_{qa}$}
\end{axis}
\end{tikzpicture}
\caption{ $\alpha$, $p_c$ and $p_{qa}$ for the context-response 
\{{\ttfamily
<CLIENT> hello brandon i was wondering how many gb s of internal memory does the boost mobil galaxy prevail have <AGENT> sorry to hear that this is giving you trouble , but i will be happy to assist you . <CLIENT> ok <AGENT> 117 mb internal storage , available to user <newline> <EOC>}\} - 
\{{\ttfamily 
plus 2 gigabyte card included <pause> <EOR>}\}. }
\label{fig alpha 2} 
\end{figure}